\documentclass{elsarticle}

\usepackage{graphicx}
\usepackage{amssymb}
\usepackage{url}
\usepackage{longtable}
\usepackage[export]{adjustbox}

\usepackage{lineno}

\journal{Computer Vision and Image Understanding}

\begin{document}
\begin{frontmatter}


\title{Systematic evaluation of CNN advances on the ImageNet}



\author[1]{Dmytro Mishkin} 
\author[2]{Nikolay Sergievskiy} 
\author[1]{Jiri Matas}
\address[1]{Center for Machine Perception, Faculty of Electrical Engineering,\\ 
Czech Technical University in Prague.  Karlovo namesti, 13. Prague 2, 12135\\
}
\address[2]{ELVEES NeoTek, Proyezd 4922, 4 build. 2, Zelenograd, Moscow,\\
Russian Federation, 124498 
}

\begin{abstract}
The paper systematically studies the impact of a range of recent advances in CNN architectures and learning methods on the object categorization (ILSVRC) problem. The evalution tests the influence of the following choices of the architecture: non-linearity (ReLU, ELU, maxout, compatability with batch normalization), pooling variants (stochastic, max, average, mixed), network width, classifier design (convolutional, fully-connected, SPP), image pre-processing, and of learning parameters:  learning rate, batch size, cleanliness of the data, etc.

The performance gains of the proposed modifications are first tested individually and then in combination. The sum of individual gains is bigger than the observed improvement when all modifications are introduced, but the "deficit" is small suggesting independence of their benefits.

We show that the use of 128x128 pixel images is sufficient to make qualitative conclusions about optimal network structure that hold for the full size Caffe and VGG nets. The results are obtained an order of magnitude faster than with the standard 224 pixel images.
\end{abstract}


\end{frontmatter}


\section{Introduction}
\label{sec:intro}
Deep convolution networks have become the mainstream method for solving various computer vision tasks, such as image classification~\cite{ILSVRC15}, object detection~\cite{ILSVRC15, PASCAL2010}, semantic segmentation~\cite{Dai2015}, image retrieval~\cite{Tolias2016}, tracking~\cite{Nam2015}, text detection~\cite{Jaderberg2014}, stereo matching~\cite{Zbontar2014}, and many other.

Besides two classic works on training neural networks -- \cite{LeCun1998} and \cite{Bengio2012}, which are still  highly relevant, there is very little guidance or theory on the plethora of design choices and hyper-parameter settings of CNNs with the consequent that researchers proceed by trial-and-error experimentation and  architecture copying, sticking to established net types.  With good results in ImageNet competition, the AlexNet~\cite{AlexNet2012},  VGGNet~\cite{VGGNet2015} and GoogLeNet(Inception)~\cite{Googlenet2015} have become the de-facto standard.

Improvements of many components of the CNN architecture like the non-linearity type, pooling, structure and learning have been recently proposed. First applied in the ILSVRC~\cite{ILSVRC15} competition, they have been adopted in different research areas.

The contributions of the recent CNN improvements and their interaction have not been systematically evaluated. We survey the recent developments and perform a large scale experimental study that considers the choice of non-linearity, pooling, learning rate policy, classifier design, network width, batch normalization~\cite{BatchNorm2015}.
We did not include ResNets~\cite{DeepResNet2015} -- a recent development achieving excellent results -- since they have been  well covered in papers~\cite{He2016,Szegedy2016,WideResNets2016,FractalNets2016}.

There are three main contributions of the paper. 
First, we survey and present baseline results for a wide variety of architectures and design choices both alone and in combination. Based on large-scale evaluation, we provide novel recommendations and insights about construction deep convolutional network.
Second, we present ImageNet-128px as fast (24 hours of training AlexNet on GTX980) and reliable benchmark -- the relative order of results for popular architectures does not change compared to common image size 224x224 or even 300x300 pixels. 
Last, but not least, the benchmark is fully reproducible and all scripts and data are available online\footnote{\url{https://github.com/ducha-aiki/caffenet-benchmark}}. 

The paper is structured as follows. In Section~\ref{sec:framework} we explain and validate experiment design. In Section~\ref{sec:single-exp}, the
influence of the a range of hyper-parameters is evaluated in isolation. The related literature is review the corresponding in experiment sections.  Section~\ref{sec:best-exp} is devoted to the combination of best hyper-parameter setting  and to ``squeezing-the-last-percentage-points'' for a given architecture recommendation. The paper is concluded in Section~\ref{sec:conclusion}.

\section{Evaluation}
Standard CaffeNet parameters and architecture are shown in Table~\ref{tab:basic-arch}. The full list of tested attributes is given in Table~\ref{tab:what-we-tested}.

\begin{table}[htb]
\caption{List of hyper-parameters tested.}
\label{tab:what-we-tested}
\centering
\setlength{\tabcolsep}{.3em}
\begin{tabular}{|l|l|}
\hline
\textbf{Hyper-parameter} & \textbf{Variants} \\
\hline
Non-linearity & linear, tanh, sigmoid, ReLU, VLReLU, RReLU,\\
&PReLU, ELU, maxout, APL, combination \\
\hline
Batch Normalization (BN) & before non-linearity. after non-linearity  \\
\hline
BN + non-linearity& linear, tanh, sigmoid, ReLU, VLReLU,\\ 
&RReLU, PReLU, ELU, maxout \\
\hline
Pooling &max, average, stochastic, max+average,\\ 
&strided convolution \\
\hline
Pooling window size &  3x3, 2x2, 3x3 with zero-padding \\
\hline
Learning rate decay policy & step, square, square root, linear \\
\hline
Colorspace \& Pre-processing & RGB, HSV, YCrCb, grayscale, learned,\\
& CLAHE, histogram equalized \\
\hline
Classifier design & pooling-FC-FC-clf, SPP-FC-FC-clf, \\
&pooling-conv-conv-clf-avepool,\\ 
&pooling-conv-conv-avepool-clf\\
\hline
Network width & 1/4, 1/$2\sqrt{2}$, 1/2, 1/$\sqrt{2}$, 1 ,$\sqrt{2}$, 2, 2$\sqrt{2}$, 4, 4$\sqrt{2}$ \\
\hline
Input image size & 64, 96, 128, 180, 224 \\
\hline
Dataset size & 200K, 400K, 600K, 800K, 1200K(full)\\
\hline
Batch size & 1, 32, 64, 128, 256, 512, 1024\\
\hline
Percentage of noisy data & 0,  5\%, 10\%, 15\%, 32\%\\
\hline
Using bias & yes/no \\
\hline
\end{tabular}
\end{table}
\subsection{Evaluation framework}
\label{sec:framework}
All tested networks were trained on the 1000 object category classification problem on the ImageNet dataset~\cite{ILSVRC15}. The set consists of a 1.2M image training set, a 50K image validation set and a 100K image test set. The test set is not used in the experiments. 
The commonly used pre-processing includes image rescaling to 256xN, where $N \geq 256$, and then cropping a random 224x224 square~\cite{AlexNet2012,Howard2013}. The setup achieves good results in classification, but training a network of this size takes several days even on modern GPUs. We thus propose to limit the image size to 144xN where $N \geq 128$ (denoted as ImageNet-128px). For example, the CaffeNet~\cite{jia2014caffe} is trained within 24 hours using NVIDIA GTX980 on ImageNet-128px. 

\begin{figure}[tb]
\centering
\includegraphics[width=0.49\linewidth]{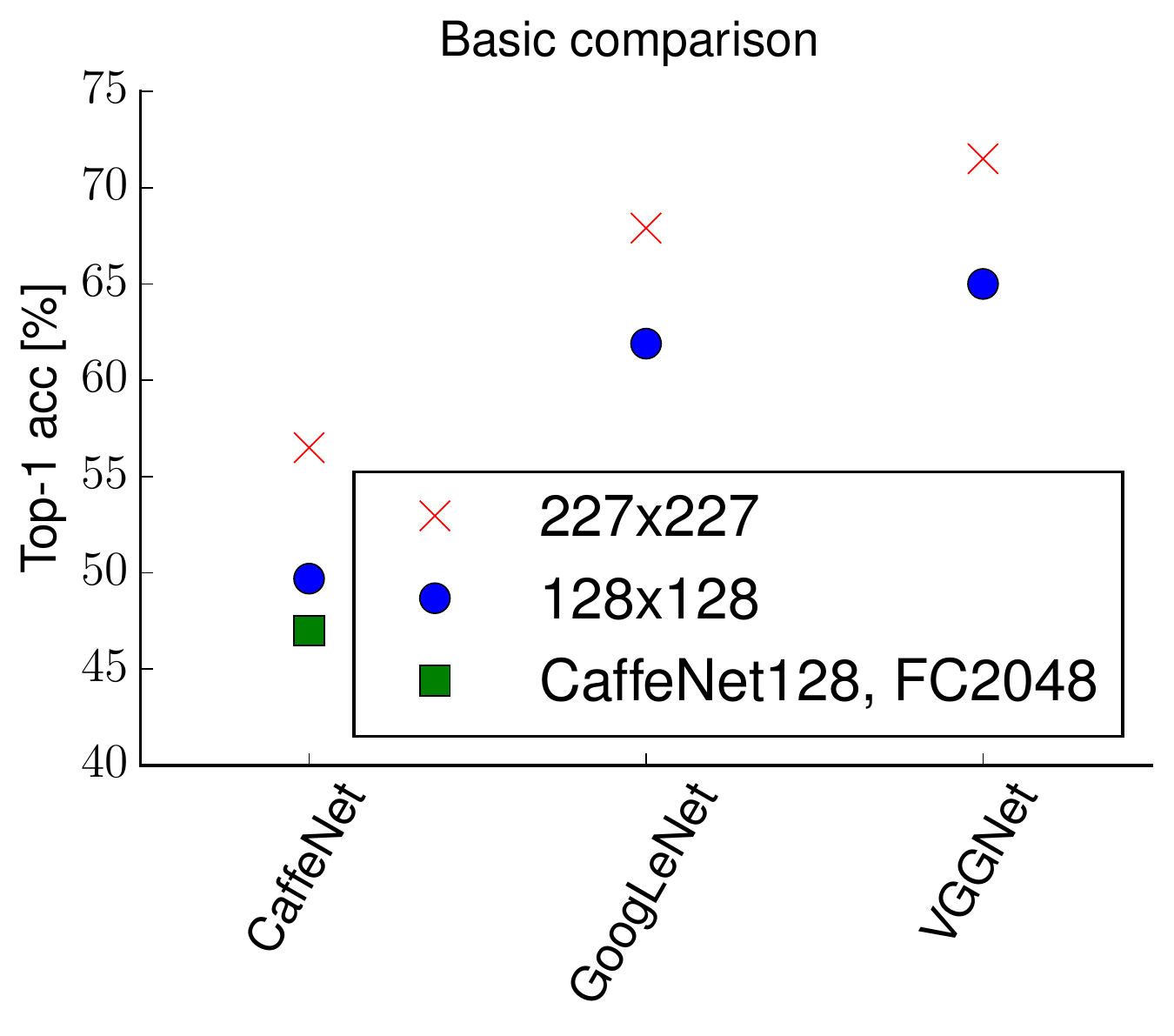}\\
\caption{Impact of image and network size on top-1 accuracy. }
\label{fig:basic-scale-validation}
\end{figure}

\subsubsection{Architectures}
The input size reduction is validated by training CaffeNet, GoogLeNet and VGGNet on both the reduced and standard image sizes. The results are shown in Figure~\ref{fig:basic-scale-validation}. The reduction of the input image size leads to a consistent drop in top-1 accuracy around 6\% for all there popular architectures and does not change their relative order (VGGNet $>$ GoogLeNet $>$ CaffeNet) or accuracy difference.  

In order to decrease the probability of overfitting and to make experiments less demanding in memory, another change of CaffeNet is made. A number of filters in fully-connected layers 6 and 7 were reduced by a factor of two, from 4096 to 2048. The results validating the
resolution reduction are presented in Figure~\ref{fig:basic-scale-validation}.

The parameters and architecture of the standard CaffeNet are shown in Table~\ref{tab:basic-arch}. For experiments we used CaffeNet with 2x thinner fully-connected layers, named as CaffeNet128-FC2048. The architecture can be denoted as 96C11/4 $\rightarrow$ MP3/2 $\rightarrow$ 192G2C5/2 $\rightarrow$ MP3/2 $\rightarrow$ 384G2C3 $\rightarrow$ 384C3 $\rightarrow$ 256G2C3 $\rightarrow$ MP3/2 $\rightarrow$ 2048C3 $\rightarrow$ 2048C1 $\rightarrow$ 1000C1. Here we used fully-convolutional notation for fully-connected layers, which are equivalent when image input size is fixed to 128x128 px. The default activation function is ReLU and it is put after every convolution layer, except the last 1000-way softmax classifier. 

\subsubsection{Learning}
SGD with momentum 0.9 is used for learning, the initial learning rate is set to 0.01, decreased by a factor of ten after each 100K iterations until learning stops after 320K iterations. The L2 weight decay for convolutional weights is set to 0.0005 and it is not applied to bias. The dropout~\cite{Dropout2014} with probability 0.5 is used before the two last layers. All the networks were initialized with LSUV~\cite{Mishkin2016LSUV}. Biases are initialized to zero. Since the LSUV initialization works under assumption of preserving unit variance of the input, pixel intensities were scaled by 0.04, after subtracting the mean of BGR pixel values (104 117 124).
\begin{table}[htb]
\caption{The basic CaffeNet architecture used in most experiments. Pad 1 -- zero-padding on the image boundary with1 pixel. Group 2 convolution -- filters are split into 2 separate groups. The architecture is denoted in ``shorthand'' as 96C11/4 $\rightarrow$ MP3/2 $\rightarrow$ 192G2C5/2 $\rightarrow$ MP3/2 $\rightarrow$ 384G2C3 $\rightarrow$ 384C3 $\rightarrow$ 256G2C3 $\rightarrow$ MP3/2 $\rightarrow$ 2048C3 $\rightarrow$ 2048C1 $\rightarrow$ 1000C1.}
\label{tab:basic-arch}
\centering
\setlength{\tabcolsep}{.3em}
\begin{tabular}{|l|c|}
\hline
input & image 128x128 px, random crop from 144xN, random mirror\\
\hline
pre-process & out = 0.04 (BGR - (104; 117; 124))\\
\hline
conv1 &conv 11x11x96, stride 4 \\
&ReLU \\
pool1&max pool 3x3, stride 2 \\
\hline
conv2&conv 5x5x256, stride 2, pad 1, group 2 \\
&ReLU \\
pool2 &max pool 3x3, stride 2 \\
\hline
conv3&conv 3x3x384, pad 1 \\
&ReLU \\
conv4&conv 3x3x384, pad 1, group 2 \\
&ReLU \\
conv5&conv 3x3x256, pad 1, group 2 \\
&ReLU \\
pool5 &max pool 3x3, stride 2 \\
\hline
fc6& fully-connected 4096 \\
&ReLU \\
drop6 & dropout ratio 0.5\\
fc7& fully-connected 4096 \\
&ReLU \\
drop7 & dropout ratio 0.5\\
\hline
fc8-clf & softmax-1000\\
\hline
\end{tabular}
\end{table}

\section{Single experiments}
\label{sec:single-exp}
This section is devoted to the experiments with a single hyper-parameter or design choice per experiment.

\subsection{Activation functions}
\subsubsection{Previous work}
The activation functions for neural networks are a hot topic, many functions have been proposed since the ReLU discovery~\cite{ReLU2011}. The first group are related to ReLU, i.e. LeakyReLU~\cite{Maas2013} and Very Leaky ReLU~\cite{GrahamCIFAR}, RReLU~\cite{RReLU2015},PReLU~\cite{PReLU2015} and its generalized version -- APL~\cite{APL2014}, ELU~\cite{ELU2016}. Others are based on different ideas, e.g. maxout~\cite{Maxout2013}, MBA~\cite{MBA2016}, etc. However, to our best knowledge only a small fraction of this activation functions have been evaluated on ImageNet-scale dataset. And when they have, e.g. ELU, the network architecture used in the evaluation was designed specifically for the experiment and is not commonly used. 

\subsubsection{Experiment}
\begin{table}[htb]
\caption{Non-linearities tested.}
\label{tab:nonlin-list}
\centering
\setlength{\tabcolsep}{.3em}
\begin{tabular}{llc}
\hline
\textbf{Name} & \textbf{Formula} & \textbf{Year} \\ 
\hline
none & y = x & - \\
sigmoid & y = $\frac{1}{1+e^{-x}}$ & 1986 \\
tanh & y = $\frac{e^{2x} - 1}{e^{2x}+1}$ & 1986 \\
ReLU & y = max(x, 0) & 2010 \\
(centered) SoftPlus & y = $\ln{(e^x+1)} - \ln{2}$ & 2011 \\
LReLU & y = max(x, $\alpha$x), $\alpha \approx 0.01$ & 2011 \\
maxout & y = max($W_1$x + $b_1$,$W_2$x + $b_2$)  & 2013 \\
APL &  y = max(x,0) + $\sum_{s=1}^{S}{a_{i}^{s}\max{(0,-x+b^{s}_{i}})}$ & 2014\\
VLReLU & y = max(x, $\alpha$x), $\alpha \in {0.1, 0.5 }$ & 2014 \\
RReLU & y = max(x, $\alpha$x), $\alpha$ = random(0.1, 0.5) & 2015 \\
PReLU & y = max(x, $\alpha$x), $\alpha$ is learnable & 2015 \\
ELU & y = x, if x $\geq$ 0, else $\alpha (e^{x} - 1)$ & 2015 \\
\hline
\end{tabular}
\end{table}
We have tested the most popular activation functions and all those with available or trivial implementations:
ReLU, tanh, sigmoid, VLReLU, RReLU, PReLU, ELU, linear, maxout, APL, SoftPlus. Formulas and references are given in Table~\ref{tab:nonlin-list}. We have selected APL and maxout with two linear pieces. Maxout is tested in two modifications: MaxW -- having the same effective network width, which doubles the number of parameters and computation costs because of the two linear pieces, and MaxS -- having same computational complexity - with $\sqrt{2}$ thinner each piece. Besides this, we have tested "optimally scaled" tanh, proposed by LeCun~\cite{LeCun1998}. We have also tried to train sigmoid~\cite{Rumelhart1986} network, but the initial loss never decreased.
Finally, as proposed by Swietojanski et.al~\cite{ConvReLUFCMaxout2014}, we have tested combination of ReLU for first layers and maxout for the last layers of the network.  

\begin{figure}[tb]
\centering
\includegraphics[width=0.80\linewidth]{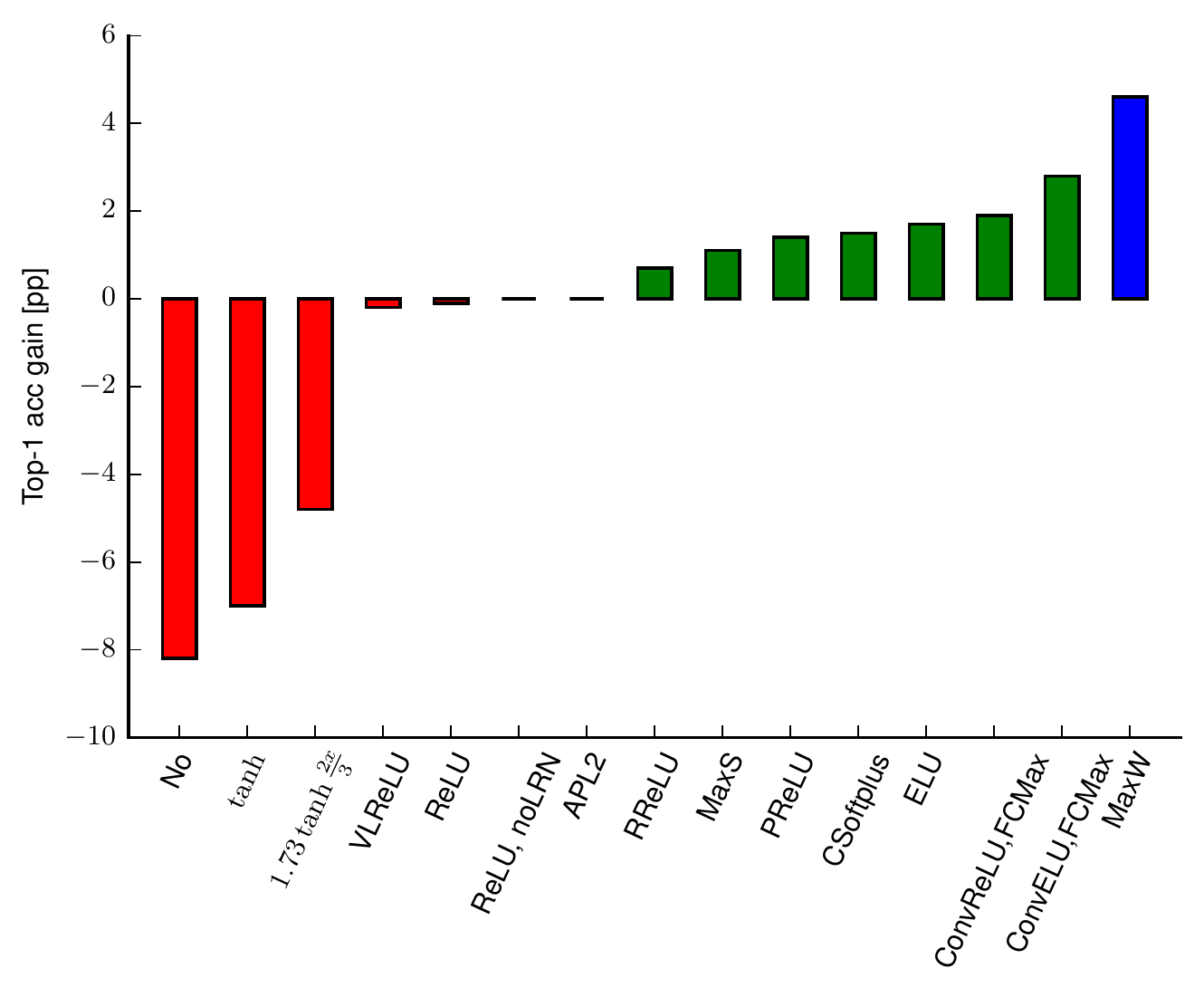}\\
\caption{Top-1 accuracy gain over ReLU in the CaffeNet-128 architecture. MaxS stands for "maxout, same compexity", MaxW -- maxout, same width, CSoftplus -- centered softplus. The baseline, i.e. ReLU, accuracy is 47.1\%.}
\label{fig:activations}
\end{figure}

Results are shown in Figure~\ref{fig:activations}.
The best single performing activation function similar in complexity to ReLU is ELU. The parametric PReLU performed on par. The performance of the centered softplus is the same as for ELU. Surprisingly, Very Leaky ReLU, popular for DCGAN networks~\cite{DCGAN2015} and for small datasets, does not outperforms vanilla ReLU. Interesting, the network with no non-linearity has respectable performance -- 38.9\% top-1 accuracy on ImageNet, not much worse than tanh-network.

The Swietojanski et.al~\cite{ConvReLUFCMaxout2014} hypothesis about maxout power in the final layers is confirmed and combined ELU (after convolutional layers) + maxout (after fully connected layers) shows the best performance among non-linearities with speed close to ReLU. Wide maxout outperforms the rest of the competitors at a higher computational cost.  

\subsection{Pooling}
\subsubsection{Previous work}

Pooling, combined with striding, is a common way to archive a degree of invariance together with a reduction of spatial size of feature maps. The most popular options are max pooling and average pooling.
Among the recent advances are: Stochastic pooling~\cite{StochasticPool2013}, LP-Norm pooling~\cite{LPNormPool2013} and Tree-Gated pooling~\cite{GenPool2015}. Only the authors of the last paper have tested their pooling on ImageNet. 

The pooling receptive field is another design choice. Krizhevskiy etal.~\cite{AlexNet2012} claimed superiority of overlapping pooling with 3x3 window size and stride 2, while VGGNet~\cite{VGGNet2015} uses a non-overlapping 2x2 window.

\subsubsection{Experiment}
\begin{table}[htb]
\caption{Poolings tested.}
\label{tab:pooling-list}
\centering
\setlength{\tabcolsep}{.3em}
\begin{tabular}{lll}
\hline
\textbf{Name} & \textbf{Formula} & \textbf{Year} \\ 
\hline
max & y = $\max_{i,j=1}^{h,w}{x_{i,j}}$ & 1989 \\ 
average & y = $\frac{1}{hw}\sum_{i,j=1}^{h,w}{x_{i,j}}$ & 1989 \\ 
stochastic & y = $x_{i,j}$ with prob. $\frac{x_{i,j}}{\sum_{i,j=1}^{h,w}{x_{i,j}}}$ & 2013 \\ 
strided convolution & \multicolumn{1}{c}{--} & 2014\\
max + average & y =$\max_{i,j=1}^{h,w}{x_{i,j}} + \frac{1}{hw}\sum_{i,j=1}^{h,w}{x_{i,j}}$ & 2015\\
\hline
\end{tabular}
\end{table}
\begin{figure}[tb]
\centering
\includegraphics[width=0.49\linewidth]{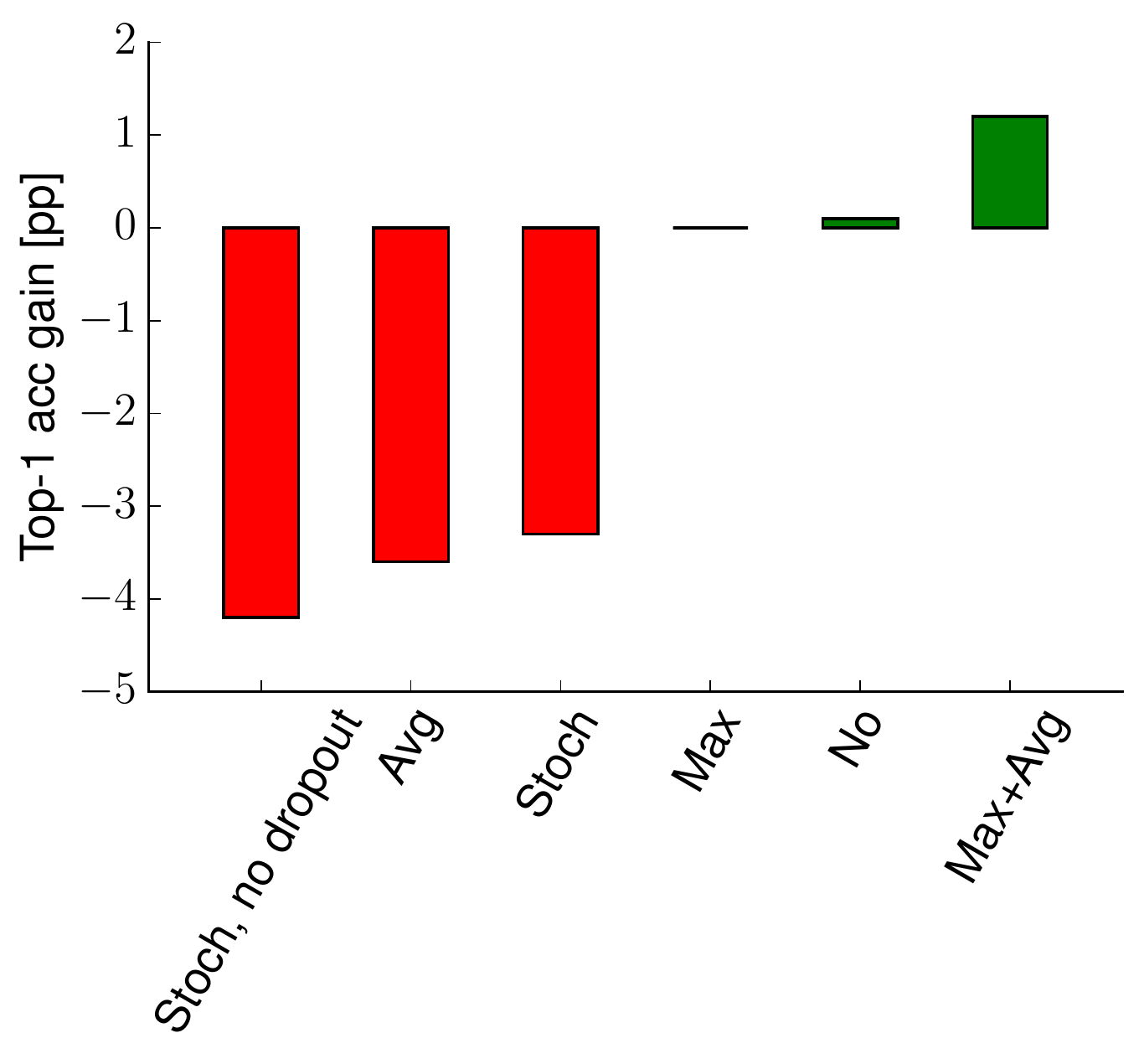}
\includegraphics[width=0.49\linewidth]{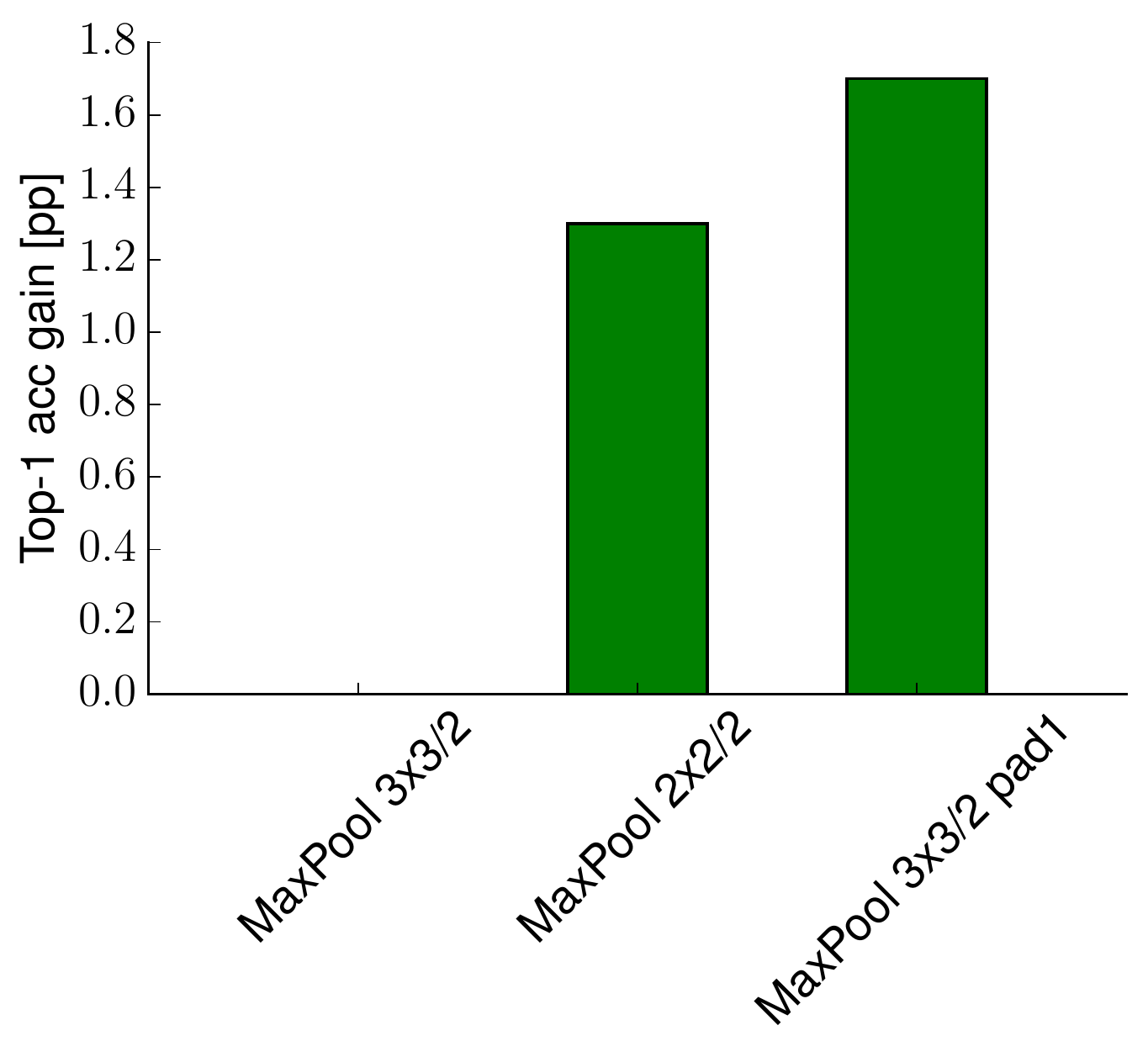}\\
\caption{Top-1 accuracy gain over max pooling for the CaffeNet-128 architecture. Left -- different pooling methods, right -- different receptive field sizes. Stoch stands for stochastic pooling, "stoch no dropout" -- for a network with stochastic pooling and turned off drop6 and drop7 layers.}
\label{fig:pooling}
\end{figure}

 We have tested (see Table~\ref{tab:pooling-list}) average, max, stochastic and proposed by Lee et al~\cite{GenPool2015} sum of average and max pooling, and skipping pooling at all, replacing it with strided convolutions proposed by Springenberd et al.~\cite{ALLCNN2015}. We have also tried Tree and Gated poolings~\cite{GenPool2015}, but we encountered convergence problems and  the results were strongly depend on the input image size. We do not know if it is a problem of the implementation, or the method itself and therefore omitted the results.  

The results are shown in Figure~\ref{fig:pooling}, left. Stochastic pooling had very bad results. In order to check if it was due to extreme randomization by the stochastic pooling and dropout, we trained network without the dropout. This decreased accuracy even more. The best results were obtained by a combination of max and average pooling. Our guess is that max pooling brings selectivity and invariance, while average pooling allows using gradients of all filters, instead of throwing away 3/4 of information as done by non-overlapping 2x2 max pooling.

The second experiment is about the receptive field size. The results are shown in Figure~\ref{fig:pooling}, right. Overlapping pooling is inferior to a non-overlapping 2x2 window, but wins if zero-padding is done. This can be explained by the fact that better results are  obtained for larger outputs; 3x3/2 pooling leads to 3x3 spatial size of pool5 feature map, 2x2/2 leads to 4x4 pool5, while 3x3/2 + 1 -- to 5x5. This observation means there is a speed -- performance trade-off.
  
\subsection{Learning rate policy}
\begin{figure}[tb]
\centering
\includegraphics[width=0.49\linewidth]{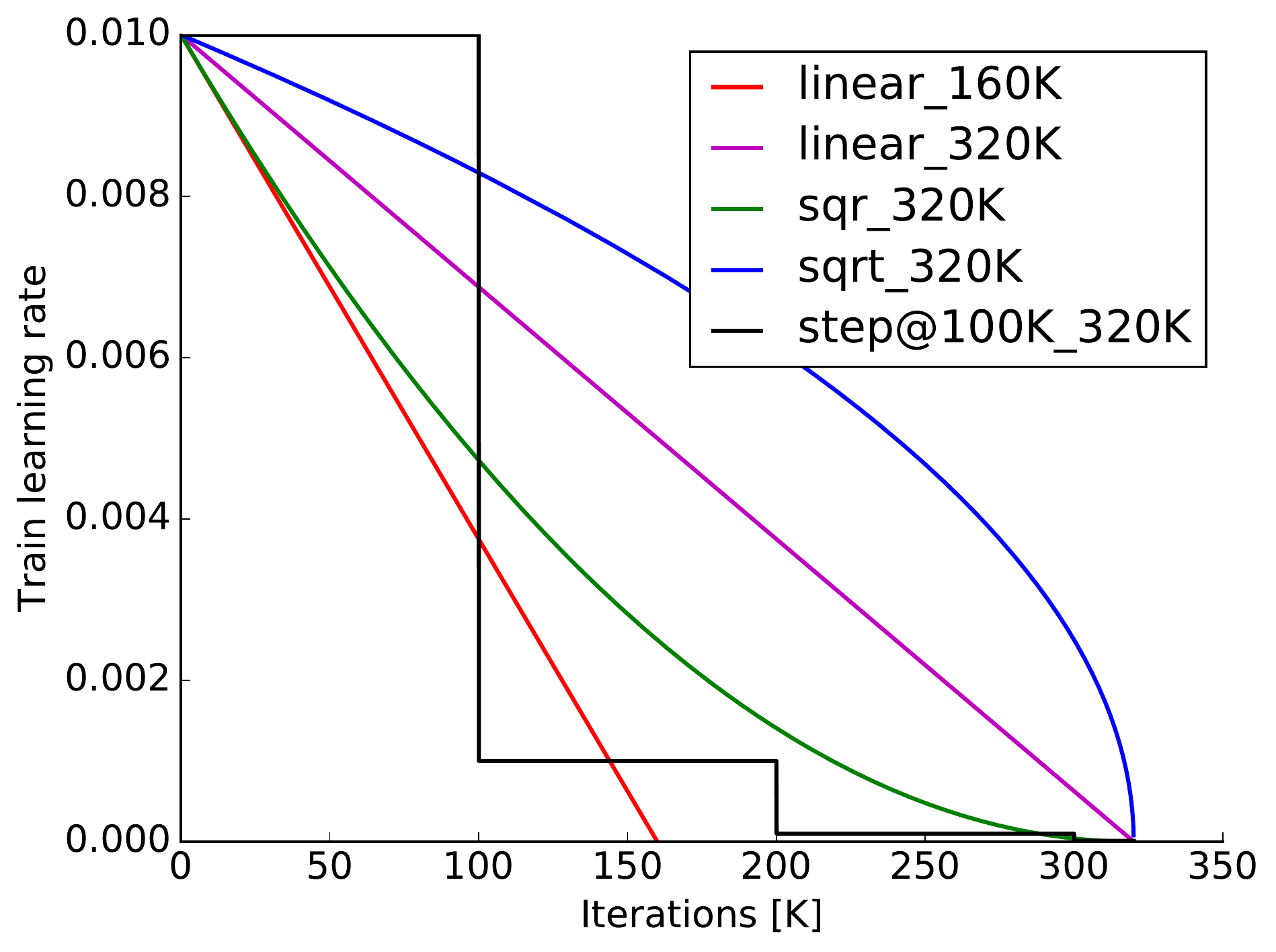}
\includegraphics[width=0.49\linewidth]{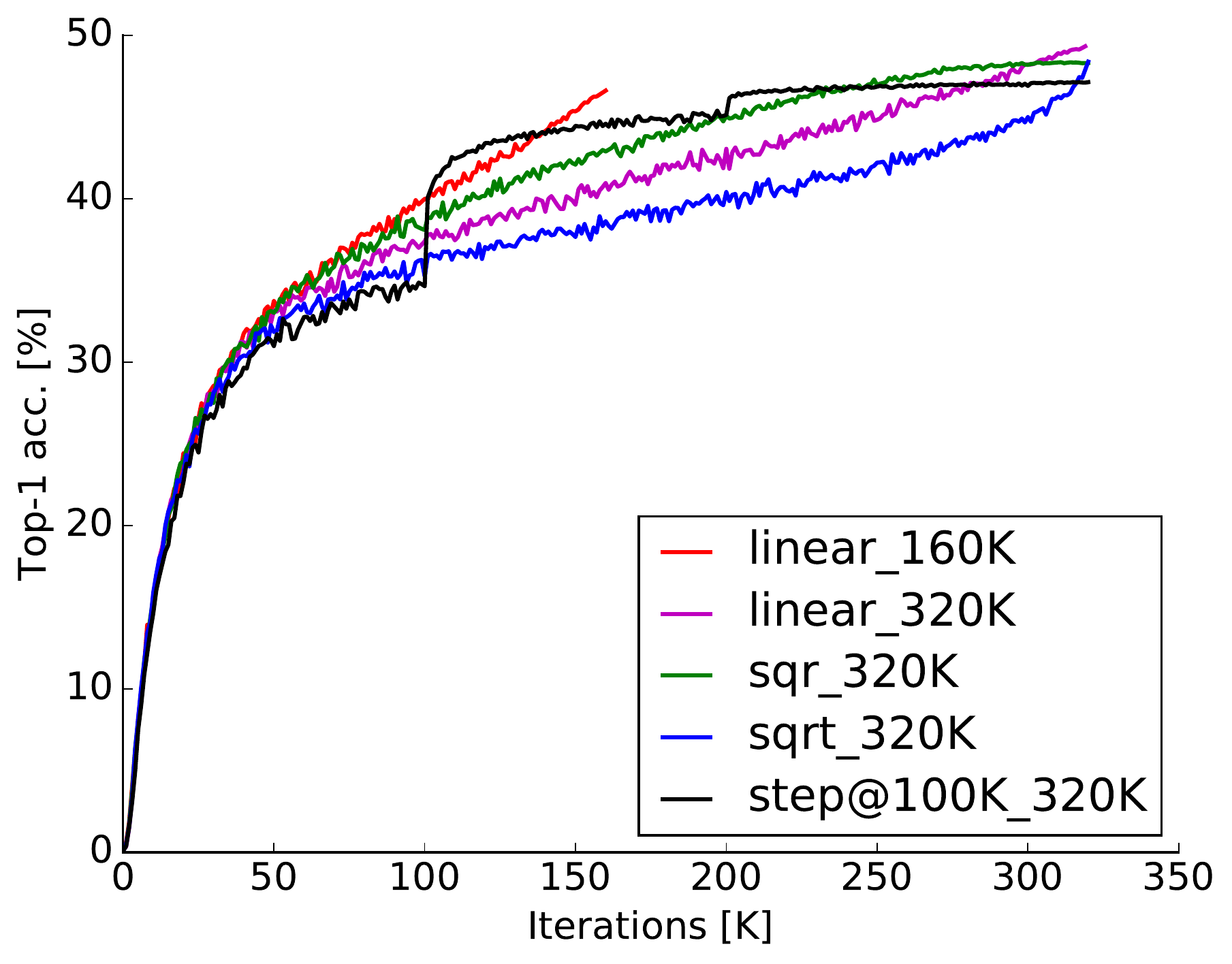}\\
\caption{Left: learning rate decay policy, right: validation accuracy. The formulas for each policy are given in Table~\ref{tab:lr-list} }
\label{fig:lr}
\end{figure}
\begin{table}[htb]
\caption{Learning rate decay policies, tested in paper.  $L_0$ -- initial learning rate, $M$ =  number of learning iterations, $i$ -- current iteration, $S$ -- step iteration. $\gamma$ -- decay coefficient .}
\label{tab:lr-list}
\centering
\setlength{\tabcolsep}{.3em}
\begin{tabular}{lllr}
\hline
\textbf{Name} & \textbf{Formula} & \textbf{Parameters} & \textbf{Accuracy} \\ 
\hline
step & lr = $L_0\gamma^{\textmd{floor(i/S)}}$ & $S$ = 100K, $\gamma=0.1$, $M=320K$ & 0.471 \\
square & lr = $L_0(1 - i/M)^2$& $M=320K$ & 0.483 \\
square root &lr = $L_0\sqrt{1 - i/M}$  & $M=320K$ & 0.483 \\
linear & lr = $L_0(1 - i/M)$& $M=320K$ & 0.493 \\
 && $M=160K$ & 0.466 \\

\hline
\end{tabular}
\end{table}
Learning rate is one of the most important hyper-parameters which influences the final CNN performance. Surprisingly, the most commonly used learning rate decay policy is "reduce learning rate 10x, when validation error stops decreasing" adopted with no parameter search. While this works well in practice, such lazy policy can be sub-optimal. 
We have tested four learning rate policies: step, quadratic and square root decay (used for training GoogLeNet by BVLC~\cite{jia2014caffe}), and linear decay. The actual learning rate dynamics are shown in Figure~\ref{fig:lr}, left. The validation accuracy is shown in the right. Linear decay gives the best results.

\subsection{Image pre-processing}
\subsubsection{Previous work}
The commonly used input to CNN is raw RGB pixels and the commonly adopted recommendation is not to use any pre-processing.  There has not been much research on the optimal colorspace or pre-processing techniques for CNN. Rachmadi and Purnama~\cite{ColorCar2015} explored different colorspaces for vehicle color identification, Dong et.al~\cite{ColorSuperRes2015} compared YCrCb and RGB channels for image super-resolution, Graham~\cite{GrahamRetinopathy2015} extracted local average color from retina images in winning solution to the Kaggle competition.  
\subsubsection{Experiment}
\begin{figure}[tb]
\centering
\includegraphics[width=0.49\linewidth,valign=t]{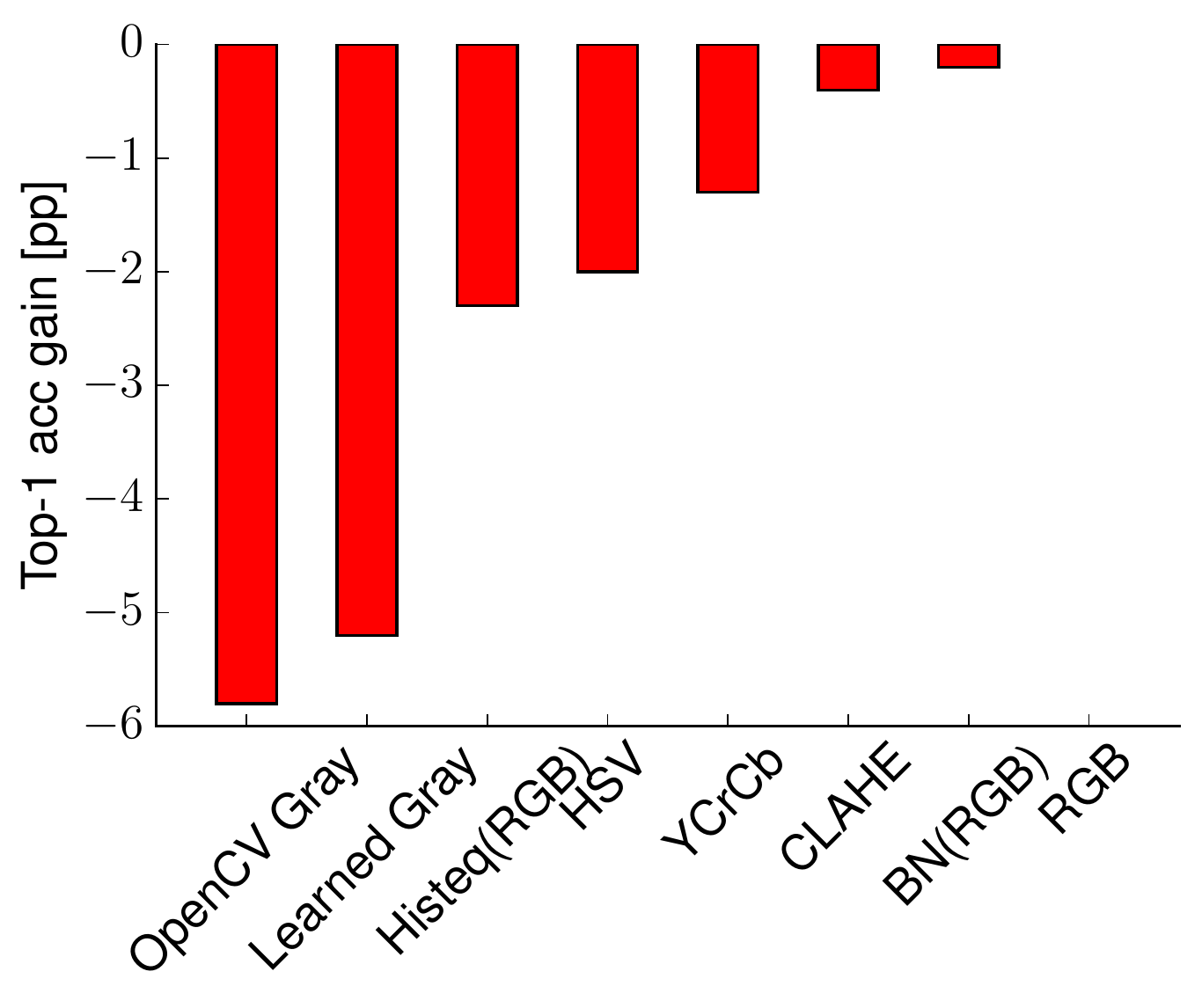}
\includegraphics[width=0.49\linewidth,valign=t]{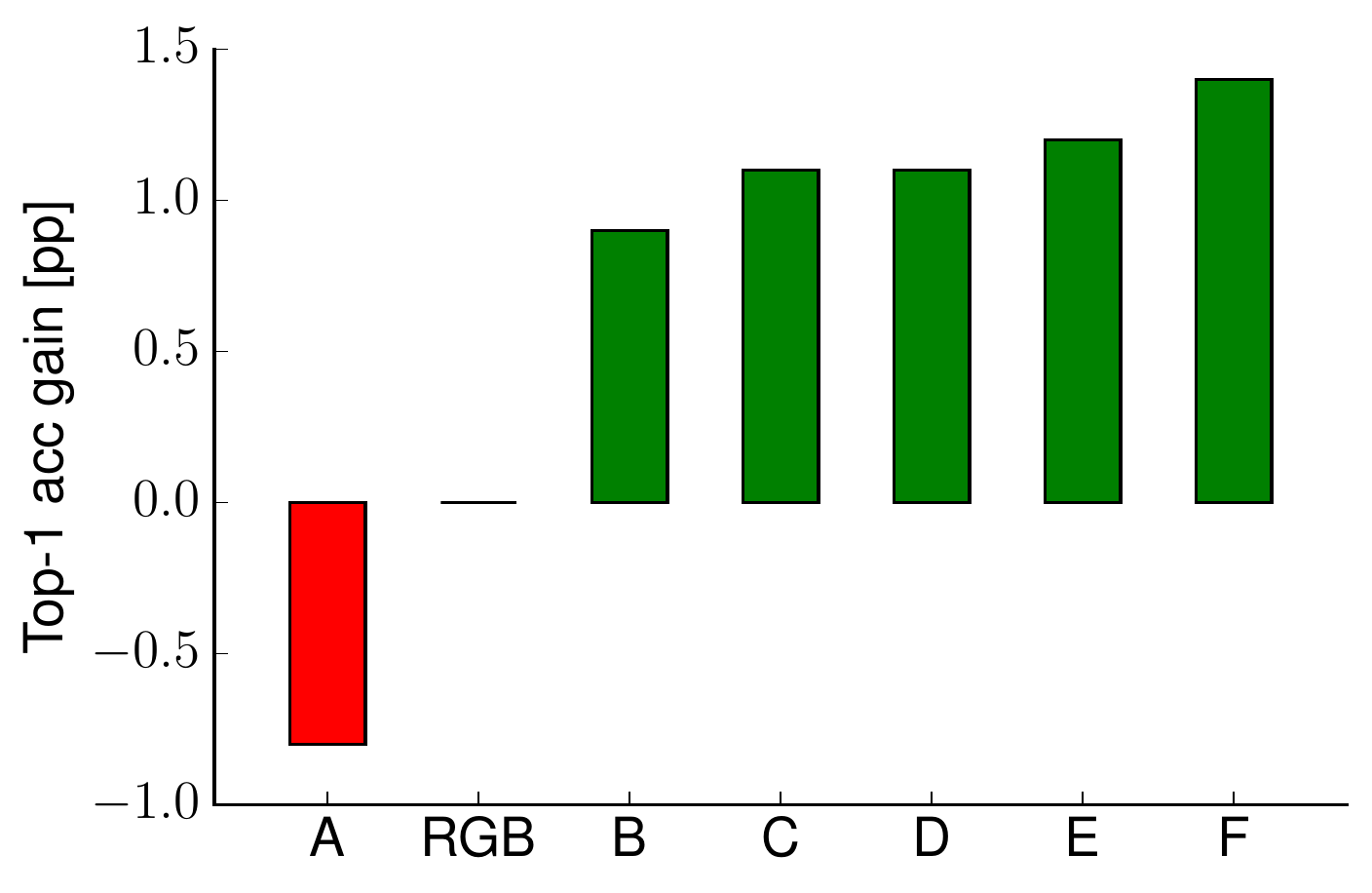}\\
\vspace{-1em}
\caption{Left: performance of using various colorspaces and pre-processing. Right: learned colorspace transformations. Parameters are given in Table~\ref{tab:colorspace-arch}.}
\label{fig:colospace}
\end{figure}
\begin{table}[htb]
\centering
\caption{Mini-networks for learned colorspace transformations, placed after image and before conv1 layer. In all cases RGB means scales and centered input  0.04 * (Img - (104, 117,124)). }
\label{tab:colorspace-arch}
\centering
\begin{tabular}{lllr}
\hline
\textbf{Name} & \textbf{Architecture} & \textbf{Non-linearity}& \textbf{Acc.}\\
\hline
A &	RGB $\rightarrow$ conv1x1x10$\rightarrow$conv1x1x3 & tanh & 0.463 \\
RGB	& RGB	& - &0.471\\
B &	RGB $\rightarrow$ conv1x1x3$\rightarrow$conv1x1x3 & VLReLU &0.480 \\
C &	RGB $\rightarrow$ conv1x1x10$\rightarrow$ conv1x1x3 + RGB & VLReLU &0.482\\
D &	 [RGB; log(RGB)] $\rightarrow$ conv1x1x10$\rightarrow$ conv1x1x3 & VLReLU&0.482 \\
E &	RGB $\rightarrow$ conv1x1x16$\rightarrow$conv1x1x3 & VLReLU & 0.483 \\
F &	RGB $\rightarrow$ conv1x1x10$\rightarrow$conv1x1x3 & VLReLU & 0.485\\
\hline
\end{tabular}
\end{table}
The pre-processing experiment is divided in two parts. First, we have tested popular handcrafted image pre-processing methods and colorspaces. Since all transformations were done on-the-fly, we first tested if calculation of the mean pixel and variance over the training set can be replaced with applying batch normalization to input images. It decreases final accuracy by 0.3\%  and can be seen as baseline for all other methods. 
We have tested HSV, YCrCb, Lab, RGB and single-channel grayscale colorspaces. Results are shown in Figure~\ref{fig:colospace}. The experiment confirms that RGB is the best suitable colorspace for CNNs. Lab-based network has not improved the initial loss after 10K iterations.
Removing color information from images costs from 5.8\% to 5.2\% of the accuracy, for OpenCV RGB2Gray and learned decolorization resp. Global~\cite{HistEq1977} and local (CLAHE~\cite{CLAHE1994}) histogram equalizations hurt performance as well.  

Second, we let the network to learn a transformation via 1x1 convolution, so no pixel neighbors are involved. The mini-networks architectures are described in Table~\ref{tab:colorspace-arch}. The learning process is joint with the main network and can be seen as extending the CaffeNet architecture with several 1x1 convolutions at the input. The best performing network gave 1.4\% absolute accuracy gain without a significant computational cost. 

\subsection{Batch normalization}
Batch normalization~\cite{BatchNorm2015} (BN) is a recent method tha  t solves the gradient exploding/vanishing problem and guarantees near-optimal learning regime for the layer following the batch normalized one. Following~\cite{Mishkin2016LSUV}, we first tested different options where to put BN -- before or after the non-linearity. Results presented in Table~\ref{tab:bn-before-or-after} are surprisingly contradictory: CaffeNet architecture prefers Conv-ReLU-BN-Conv, while GoogLeNet -- Conv-BN-ReLU-Conv placement. Moreover, results for GoogLeNet are inferior to the plain network. The difference to~\cite{BatchNorm2015} is that we have not changed any other parameters except using BN, while in the original paper, authors decreased regularization (both weight decay and dropout), changed the learning rate decay policy and applied an additional training set re-shuffling. Also, GoogLeNet behavior seems different to CaffeNet and VGGNet w.r.t. to other modification, see Section~\ref{sec:best-exp}.

\begin{table}[htb]
\centering
\caption{Top-1 accuracy on ImageNet-128px, batch normalization placement. ReLU activation is used.}
\label{tab:bn-before-or-after}
\centering
\begin{tabular}{lrrr}
\hline
Network & \multicolumn{3}{c}{BN placement}\\
&No BN & Before & After\\
\hline
CaffeNet128-FC2048 & 0.471& 0.478& \textbf{0.499}\\
GoogLeNet128 & \textbf{0.619} & 0.603& 0.596 \\
\hline
\end{tabular}
\end{table}

\begin{figure}[tb]
\centering
\includegraphics[width=0.80\linewidth]{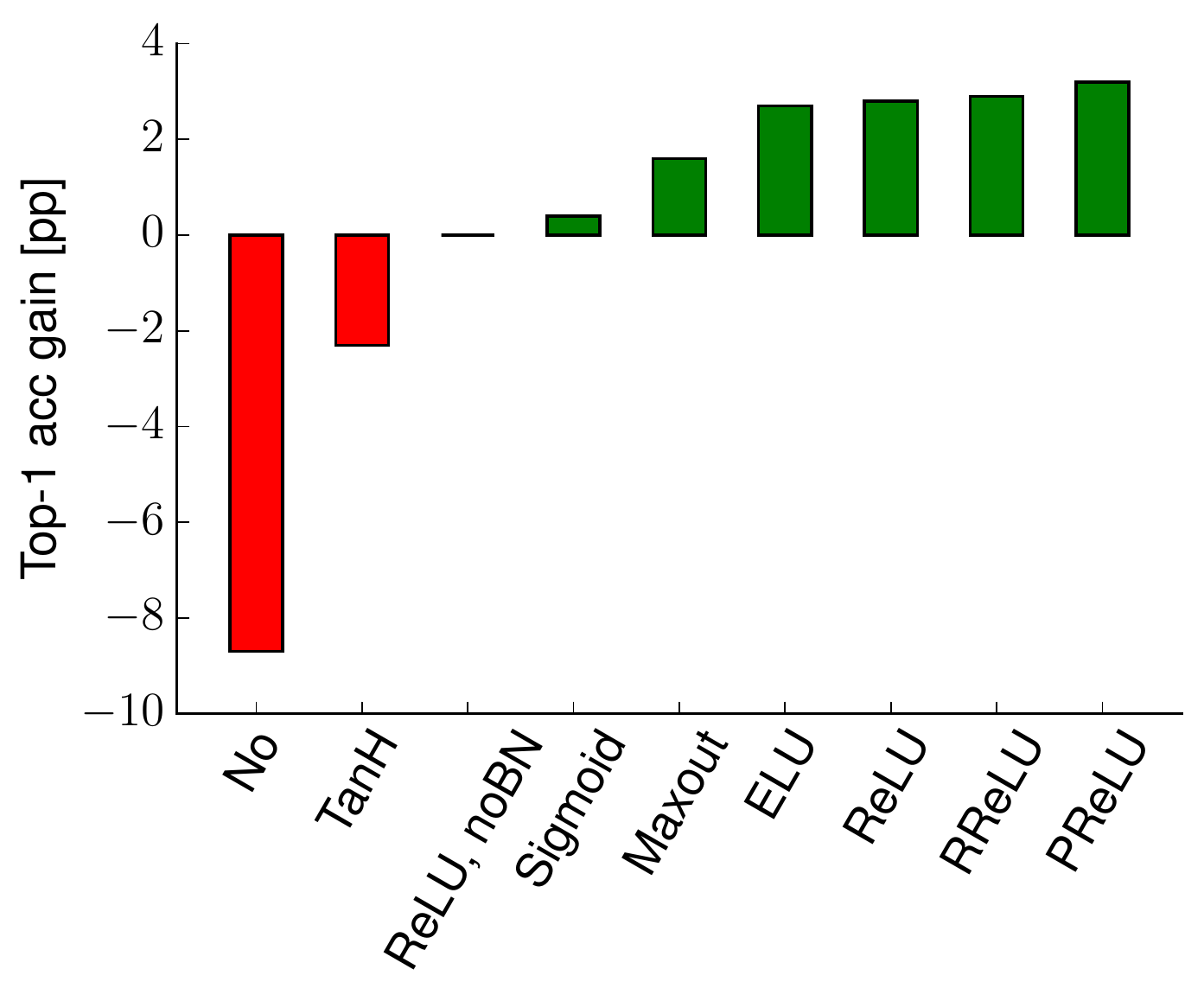}\\
\caption{Top-1 accuracy gain over ReLU without batch normalization (BN) in CaffeNet-128 architecture. The baseline -- ReLU -- accuracy is 47.1\%.}
\label{fig:bn-activations}
\end{figure}

For the next experiment with BN and activations, we selected placement after non-linearity. Results are shown in Figure~\ref{fig:bn-activations}. Batch normalization washes out differences between ReLU-family variants, so there is no need to use the  more complex variants. Sigmoid with BN outperforms ReLU without it, but, surprisingly, tanh with BN shows worse accuracy than sigmoid with BN. 

\subsection{Classifier design}
\subsubsection{Previous work}
The CNN architecture can be seen as integration of feature detector and which is following by a classifier. Ren et. al.~\cite{NoC2015} proposed to consider convolutional layers of the AlexNet as an feature extractor and fully-connected layers as 2-layer MLP as a classifier. They argued that 2 fully-connected layers are not the optimal design and explored various architectures instead. But they considered only pre-trained CNN or HOGs as feature extractor, so explored mostly transfer learning scenario, when the most of the network weights are frozen. Also, they explored architectures with additional convolution layers, which can be seen not as better classifier, but as an enhancement of the feature extractor. 

There is three the most popular approaches to classifier design. First -- final layer of the feature extractor is max pooling layer and the classifier is a one or two layer MLP, as it is done in LeNet~\cite{LeNet1998}, AlexNet~\cite{AlexNet2012} and VGGNet~\cite{VGGNet2015}. Second -- spatial pooling pyramid layer~\cite{SPPNet2014} instead of pooling layer, followed by two layer MLP. And the third architecture consist of average pooling layer, squashing spatial dimensions, followed by softmax classifier without any feature transform. This variant is used in GoogLeNet~\cite{Googlenet2015} and ResNet~\cite{DeepResNet2015}.
\subsubsection{Experiment}
\begin{figure}[tb]
\centering
\includegraphics[width=0.80\linewidth]{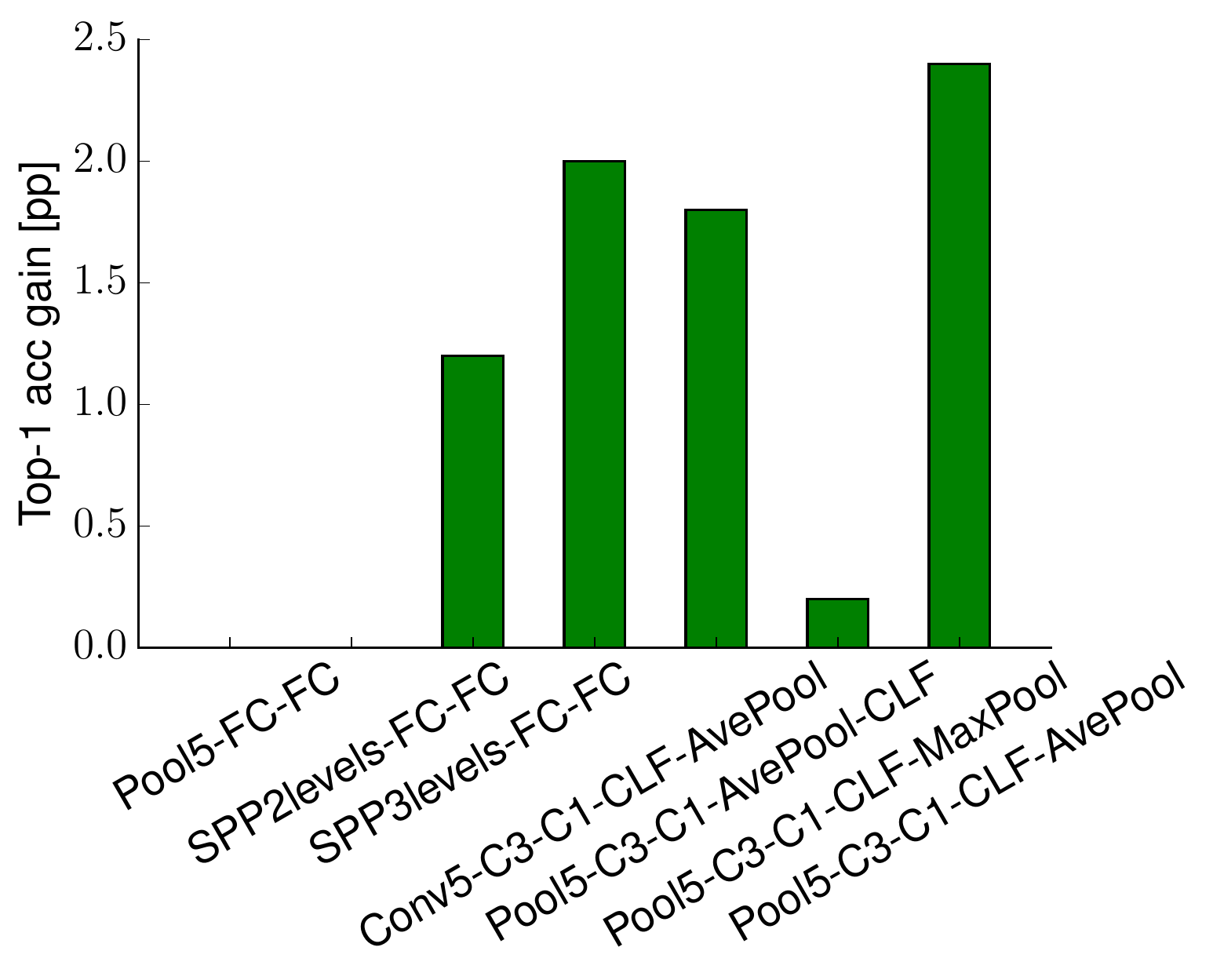}\\
\caption{Classifier design:  Top-1 accuracy gain over standard CaffeNet-128 architecture. }
\label{fig:clf-design}
\end{figure}

We have explored following variants: default 2-layer MLP, SPPNet with 2 and 3 pyramid levels, removing pool5-layer, treating fully-connected layers as convolutional, which allows to use zero-padding, therefore increase effective number of training examples for this layer, averaging features before softmax layer or averaging spatial predictions of the softmax layer~\cite{NiN2013}. The results are shown in the Figure~\ref{fig:clf-design}. The best results are get, when predictions are averaged over all spatial positions and MLP layers are treated as convolution - with zero padding. The advantage of the SPP over standard max pooling is less pronounced. 

\subsection{Batch size and learning rate}
\begin{figure}[tb]
\centering
\includegraphics[width=0.49\linewidth]{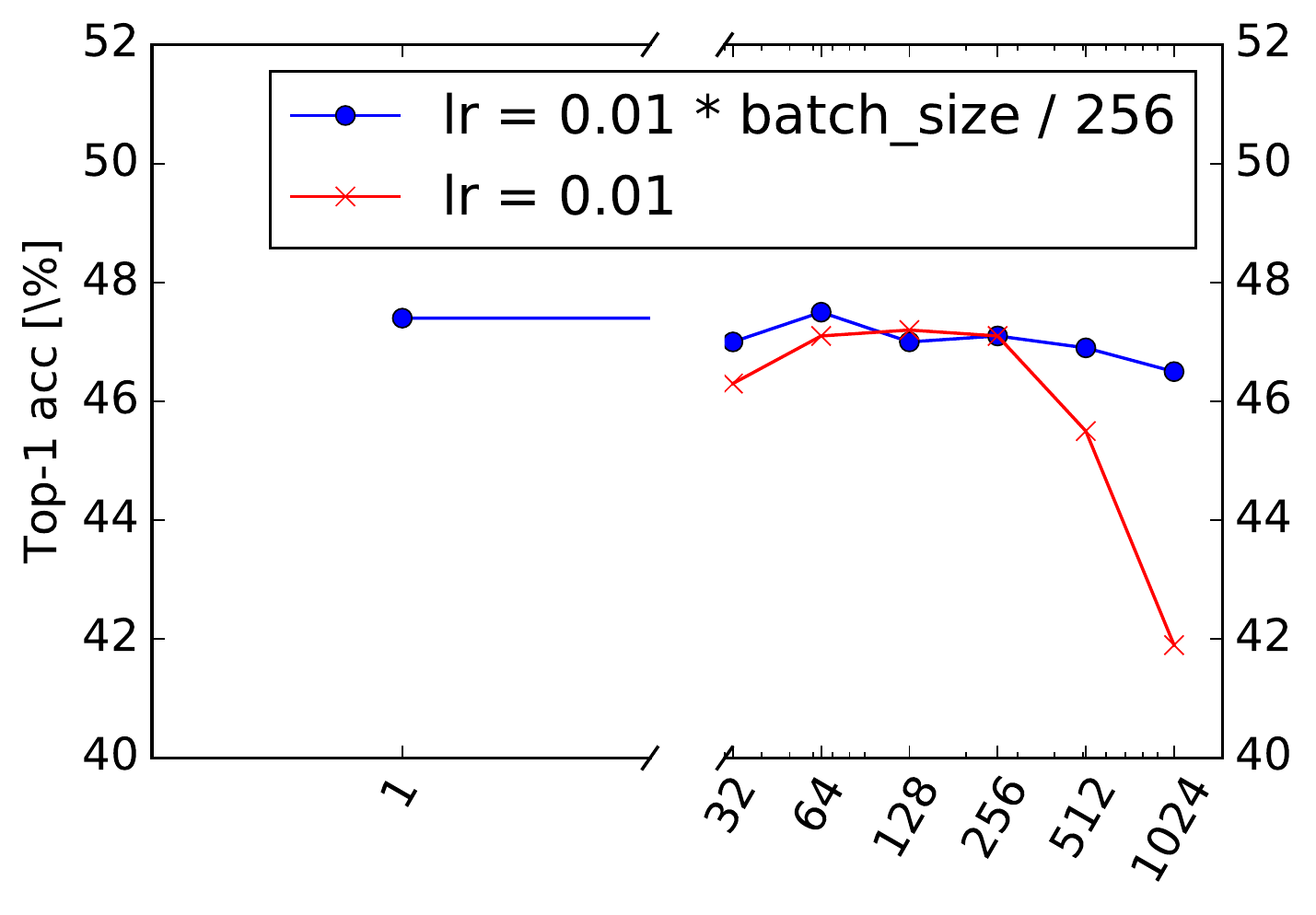}\\
\caption{Batch size and initial learning rate impact to the accuracy}
\label{fig:batchsize}
\end{figure}
The mini-batch size is always a trade-off between computation efficiency -- because GPU architecture prefers it large enough -- and accuracy; early work by Wilson and Martinez~\cite{BatchSize2003} shows superiority of the online training to batch-training. 
Here we explore the influence of mini-batch size on the final accuracy. Experiments show that keeping a constant learning rate for different mini-batch sizes has a negative impact on performance. We also have tested the heuristic proposed by Krizhevskiy~\cite{OneWeirdTrick2014} which suggests to keep the product of mini-batch size and learning rate constant. 
Results are shown in Figure\ref{fig:batchsize}. The heuristics works, but large (512 and more) mini-batch sizes leads to quite significant decrease in performance. On the other extreme, online training (mini-batch with single example) does not bring accuracy gains over 64 or 256, but significantly slows down the training wall-clock time.

\subsection{Network width}
\begin{figure}[tb]
\centering
\includegraphics[width=0.49\linewidth]{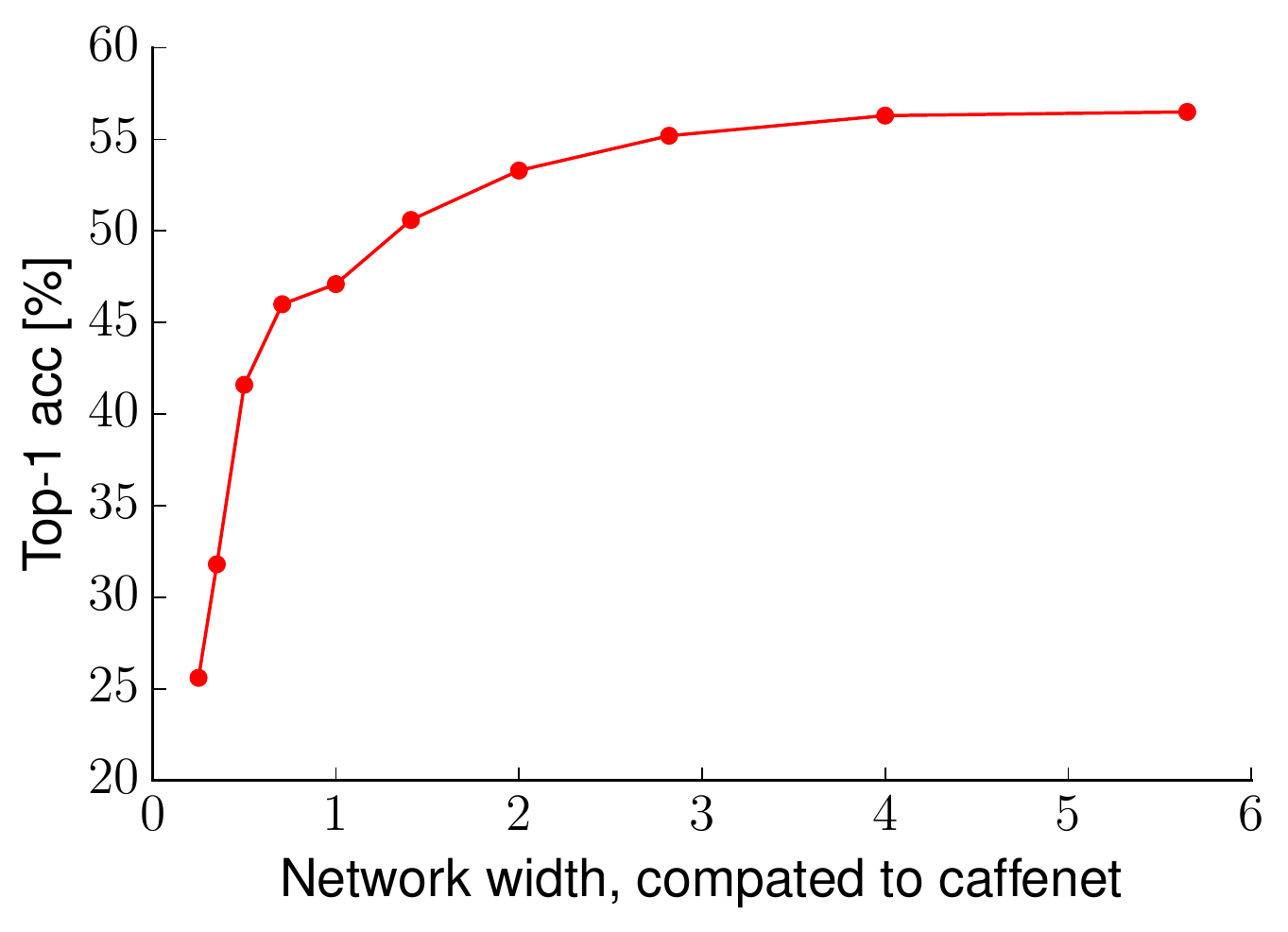}\\
\caption{Network width impact on the accuracy.}
\label{fig:network-width}
\end{figure}

All the advances in ImageNet competition so far were caused by architectural improvement.  To the best of our knowledge, there is no study about network width -- final accuracy dependence. Canziani et.al~\cite{CNNEfficiency2016} did a comparative analysis of the ImageNet winner in terms of accuracy, number of parameters and computational complexity, but it is a comparison of the different architectures. In this subsection we evaluate how far one can get by increasing CaffeNet width, with no other changes. The results are shown in Figure~\ref{fig:network-width}. The original architecture is close to optimal in accuracy per FLOPS sense: a decrease in the number of filters leads to  a quick and significant accuracy drop, while making the network thicker brings gains, but it saturates quickly. Making the network thicker more than 3 times leads to a very limited accuracy gain. 

\subsection{Input image size}
The input image size, as it brings additional information and training samples for convolution filters, plays a very important role. Our initial experiment, showed in Figure~\ref{fig:basic-scale-validation} indicates that CaffeNet, trained on 227x227 images can compete with much more complex GoogLeNet architecture, trained on smaller images. So the obvious question is what is the dependence between image size and final accuracy. 

\begin{figure}[tb]
\centering
\includegraphics[width=0.49\linewidth]{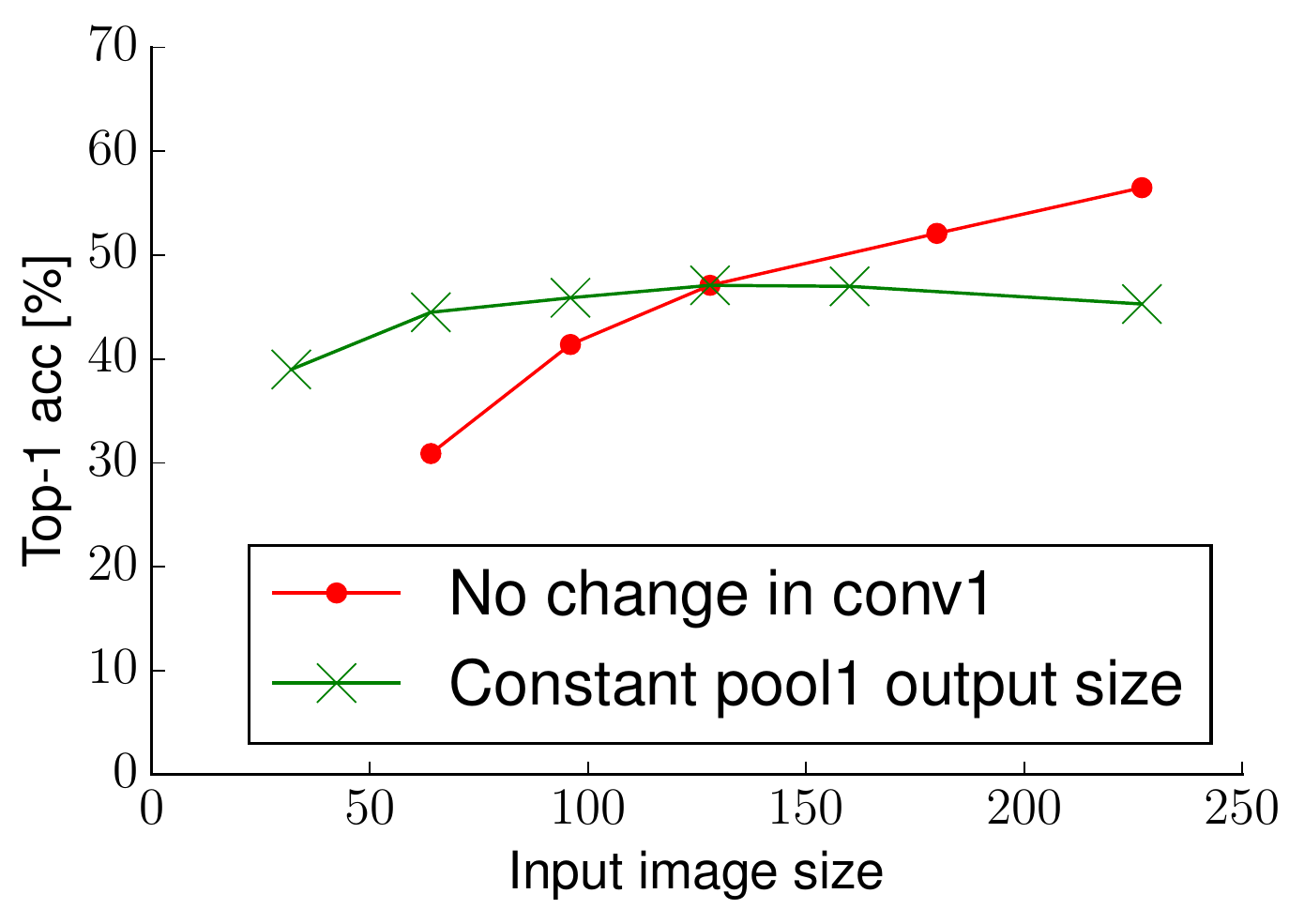}\\
\caption{Input image size impact on the accuracy}
\label{fig:input-image-size}
\end{figure}

We have performed an experiment with different input image sizes: 96, 128, 180 and 224 pixels wide. The results are presented in Figure~\ref{fig:input-image-size}. The bad news are that while accuracy depends on image size linearly, the needed computations grow quadratically, so it is a very expensive way to a performance gain. 
In the second part of experiment, we kept the spatial output size of the pool1 layer fixed while changing the input image size. To archieve this, we respectively change the stride and filter size of the conv1 layer. Results show that the gain from a large image size mostly (after some minimum value) comes from the larger spatial size of deeper layers than from the unseen image details. 
\subsection{Dataset size and noisy labels}

\begin{figure}[tb]
\centering
\includegraphics[width=0.49\linewidth]{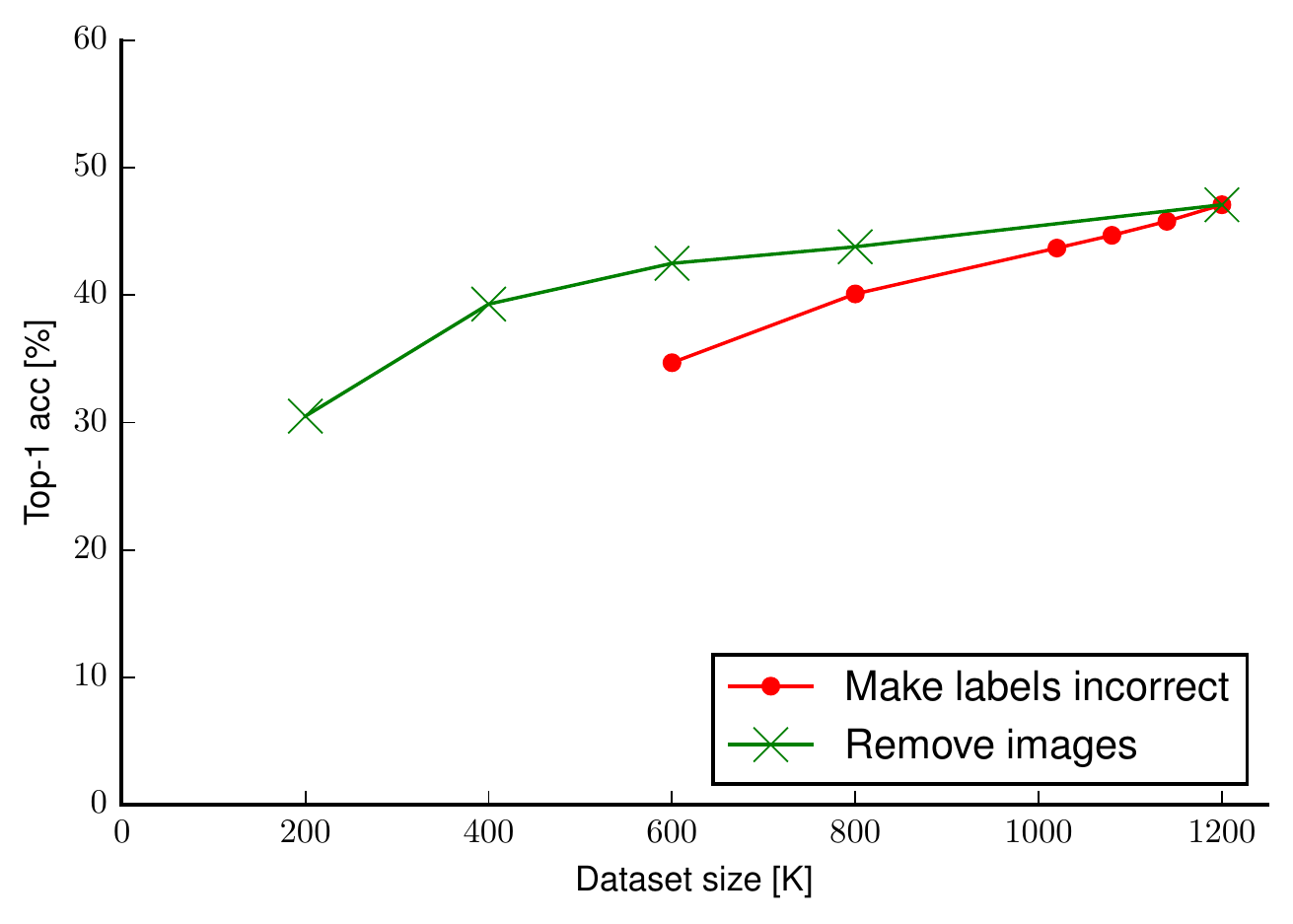}\\
\caption{Training dataset size and cleanliness impact on the accuracy}
\label{fig:dataset-size}
\end{figure}
\subsubsection{Previous work}
The performance of the current deep neural network is highly dependent on the dataset size.  Unfortunately, not much research has been published on this topic. In DeepFace~\cite{taigman2014deepface}, the  authors shows that dataset reduction from 4.4M to 1.5M leads to a 1.74\% accuracy drop. Similar dependence is shown by Schroff et.al~\cite{schroff2015facenet} but on an extra-large dataset: decreasing the dataset size from 260M to 2.6M leads to accuracy drop in 10\%. But these datasets are private and the experiments are not reproducible. 
Another important property of a dataset is the cleanliness of the data. For example, an estimate of human accuracy on ImageNet is 5.1\% for top-5~\cite{ILSVRC15}. To create the  ImageNet, each image was voted on by ten different people~\cite{ILSVRC15}. 

\subsubsection{Experiment}

We explore the dependency between the accuracy and the dataset size/cleanliness on ImageNet. For the dataset size experiment, 200, 400, 600, 800 thousand examples were random chosen from a full training set. For each reduced dataset, a CaffeNet is trained from scratch. 
 For the cleanliness test, we replaced the labes to a random incorrect one for 5\%, 10\%, 15\% and 32\% of the examples. The labels are fixed, unlike the recent work on disturbing labels as a regularization method~\cite{DisturbLabel2016}. 

The results are shown in Figure \ref{fig:dataset-size} which clearly shows that bigger (and more diverse) dataset brings an improvement. There is a minimum size below which performance quickly degrades. Less clean data outperforms more noisy ones: a clean dataset with 400K images performs on par with 1.2M dataset with 800K correct images.

\subsection{Bias in convolution layers}
We conducted a simple experiment on the importance of the bias in the convolution and fully-connected layers. First, the  network is trained as usual, for the second -- biases are initialized with zeros and the bias learning rate is set to zero. The network without biases shows 2.6\% less accuracy than the default -- see Table~\ref{tab:bias}. 
\begin{table}[htb]
\centering
\caption{Influence of the bias in convolution and fully-connected layers. Top-1 accuracy on ImageNet-128px.}
\label{tab:bias}
\centering
\begin{tabular}{lr}
\hline
\textbf{Network} & \textbf{Accuracy}\\
\hline
With bias & \textbf{0.471}\\
Without bias & 0.445\\
\hline
\end{tabular}
\end{table}

\section{Best-of-all experiments}
\label{sec:best-exp}

Finally, we test how all the improvements, which do not increase the computational cost, perform together. We combine: the learned colorspace transform F, ELU as non-linearity for convolution layers and maxout for fully-connected layers, linear learning rate decay policy, average plus max pooling. The improvements are applied to CaffeNet128, CaffeNet224, VGGNet128 and GoogleNet128. 

\begin{figure}[tb]
\centering
\includegraphics[width=0.49\linewidth]{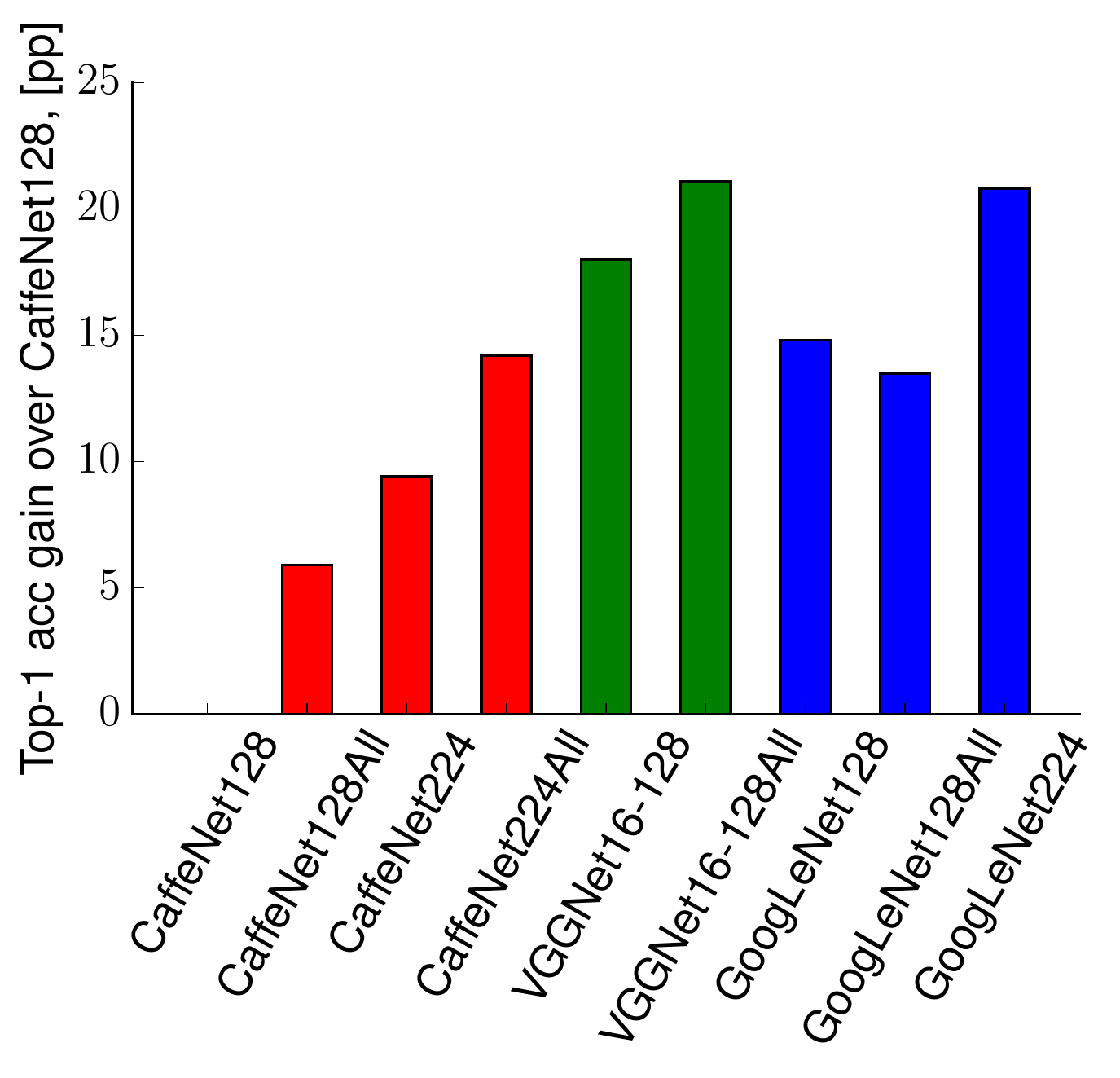}\\
\caption{Applying all improvements that do not change feature maps size: linear learning rate decay policy, colorspace transformation "F", ELU nonlinearity in convolution layers, maxout non-linearity in fully-connected layers and a sum of average and max pooling. }
\label{fig:final-test}
\end{figure}

The first three demonstrated consistent performance growth (see Figure~\ref{fig:final-test}), while GoogLeNet performance degraded, as it was found for batch normalization. Possibly, this is due to the complex and optimized structure of the GoogLeNet network.  
Unfortunately, the cost of training VGGNet224 is prohibitive, one month of GPU time, so we have not subjected it to the tests yet.

\section{Conclusions}
\label{sec:conclusion}
We have compared systematically a set of recent CNN advances on large scale dataset. We have shown that benchmarking can be done at an affordable time and computation cost. 
A summary of recommendations:
\begin{itemize}
\setlength\itemsep{-1ex}
\item use ELU non-linearity without batchnorm or ReLU with it.
\item apply a learned colorspace transformation of RGB. 
\item use the linear learning rate decay policy.
\item use a sum of the average and max pooling layers. 
\item use mini-batch size around 128 or 256. If this is too big for your GPU, decrease the learning rate proportionally to the batch size.
\item use fully-connected layers as convolutional and average the predictions for the final decision.
\item when investing in increasing training set size, check if a plateau has not been reach.
\item cleanliness of the data is more important then the size.
\item if you cannot increase the input image size, reduce the stride in the consequent layers, it has roughly the same effect.
\item if your network has a complex and highly optimized architecture, like e.g. GoogLeNet, be careful with modifications.
\end{itemize}

\begin{table}[htp]
\vspace{-13em}
\centering
\caption{Results of all tests on ImageNet-128px}
\label{tab:tbl-all}
\centering
\scriptsize
\begin{longtable}{llr}
\hline
Group Name & Names & acc [\%]\\
\hline
Baseline & & 47.1 \\
\hline
Non-linearity & Linear & 38.9 \\
&tanh & 40.1 \\
&VReLU & 46.9 \\
&APL2 & 47.1 \\
&ReLU & 47.1 \\
&RReLU & 47.8 \\
&maxout (MaxS) & 48.2 \\
&PReLU & 48.5 \\
&ELU & 48.8 \\
&maxout (MaxW) & 51.7 \\
\hline
Batch Normalization (BN) &before non-linearity & 47.4 \\
&after non-linearity & 49.9 \\
\hline
BN + Non-linearity & Linear & 38.4 \\
&tanh & 44.8 \\
&sigmoid & 47.5 \\
&maxout (MaxS) & 48.7 \\
&ELU & 49.8 \\
&ReLU & 49.9 \\
&RReLU & 50.0 \\
&PReLU &50.3 \\
\hline
Pooling &stochastic, no dropout & 42.9 \\
&average & 43.5 \\
&stochastic & 43.8 \\
&Max & 47.1 \\
&strided convolution & 47.2 \\
& max+average & 48.3 \\
\hline
Pooling window size &  3x3/2 & 47.1 \\
& 2x2/2 & 48.4 \\
&3x3/2 pad=1 & 48.8 \\
\hline
Learning rate decay policy & step & 47.1\\
&square & 48.3\\
&square root & 48.3\\
&linear & 49.3\\ 
\hline
Colorspace \& Pre-processing & OpenCV grayscale & 41.3 \\
&grayscale learned  &  41.9\\
&histogram equalized & 44.8\\
&HSV & 45.1\\
&YCrCb & 45.8 \\
&CLAHE &  46.7 \\
 & RGB &47.1 \\
\hline
Classifier design & pooling-FC-FC-clf & 47.1\\
&SPP2-FC-FC-clf & 47.1 \\
&pooling-C3-C1-clf-maxpool &  47.3 \\ 
&SPP3-FC-FC-clf & 48.3 \\
&pooling-C3-C1-avepool-clf &  48.9 \\
&C3-C1-clf-avepool &  49.1 \\
&pooling-C3-C1-clf-avepool &  49.5 \\ 
\hline
Percentage of noisy data & 5\%  & 45.8 \\
& 10\% & 44.7 \\
& 15\% & 43.7 \\
& 32\% & 40.1 \\
\hline
Dataset size & 1200K & 47.1\\  
& 800K  & 43.8 \\
& 600K & 42.5 \\
& 400K & 39.3 \\
& 200K & 30.5 \\
\hline
Network width &4$\sqrt{2}$ 	&56.5 \\
&4 	&56.3 \\
&2$\sqrt{2}$ 	&55.2 	\\
&2 	&53.3 \\
&$\sqrt{2}$ 	&50.6 \\ 
&1 &47.1 \\
&1/$\sqrt{2}$ &46.0 \\ 
&1/2 	&41.6 \\
&1/$2\sqrt{2}$ 	&31.8 \\
&1/4	&25.6\\
\hline
Batch size & BS=1024, lr=0.04 &	46.5 \\
&BS=1024, lr=0.01 	&41.9 \\
&BS=512, lr=0.02 	&46.9 \\
&BS=512, lr=0.01 	&45.5 \\
&BS=256, lr=0.01	&47.0 \\
&BS=128, lr=0.005	&47.0 \\
&BS=128, lr=0.01	&47.2 \\
&BS=64, lr=0.0025 	&47.5 \\
&BS=64, lr=0.01	&47.1 \\
&BS=32, lr=0.00125 	&47.0 \\
&BS=32, lr=0.01	&46.3 \\
&BS=1, lr=0.000039	&47.4 \\
\hline
Bias & without &	44.5 \\
&with 	&47.1 \\
\hline
Architectures & CaffeNet128 &	47.1 \\
&CaffeNet128All 	&53.0 \\
&CaffeNet224 	&56.5 \\
&CaffeNet224All 	&61.3 \\
&VGGNet16-128&	65.1\\
&VGGNet16-128All&	68.2\\
&GoogLeNet128 & 61.9	\\
&GoogLeNet128All & 60.6	\\
&GoogLeNet224 & 67.9	\\
\hline
\end{longtable}
\end{table}
\section*{Acknowledgements}
\label{sec:acknowledgement}
The authors were supported by The Czech Science Foundation Project GACR P103/12/G084 and CTU student grant 
SGS15/155/OHK3/2T/13.
\clearpage
\section{References}
\bibliographystyle{elsarticle-num}
\bibliography{main.bib}
\end{document}